\pdfoutput=1
\documentclass[11pt]{article}

\usepackage{ACL2023}
\usepackage{times}
\usepackage{latexsym}

\usepackage[T1]{fontenc}
\usepackage[utf8]{inputenc}

\usepackage{microtype}

\usepackage{inconsolata}
\usepackage{bbm}
\usepackage{amsmath}
\usepackage{makecell}
\usepackage{array,booktabs}
\newcommand {\otoprule}{\midrule [\heavyrulewidth]}
\newcolumntype {+}{ >{\global\let\currentrowstyle\relax}}
\newcolumntype {^}{ >{\currentrowstyle }}
 \newcommand {\rowstyle}[1]{\gdef\currentrowstyle{#1} %
 #1\ignorespaces
 }
\newcommand{\tabhead}{\rowstyle{\bfseries}}

\usepackage{tabularx}
\usepackage{pifont}
\newcommand{\redcross}{\textcolor{red}{\ding{55}}}
\newcommand{\greencheck}{\textcolor{green}{\ding{51}}}

\usepackage{xcolor}
\usepackage{graphicx} 
\usepackage{soul}
\usepackage{array} % Add this to your preamble

\newcolumntype{P}[1]{>{\centering\arraybackslash}p{#1}}

\definecolor{pastelblue}{HTML}{A1C9F4}
\definecolor{pastelorange}{HTML}{FFB482}
\definecolor{pastelgreen}{HTML}{8DE5A1}
\definecolor{pastelred}{HTML}{FF9F9B}
\definecolor{pastelpurple}{HTML}{D0BBFF}

\colorlet{pastelbluemuted}{pastelblue!75}
\colorlet{pastelorangemuted}{pastelorange!75}
\colorlet{pastelgreenmuted}{pastelgreen!75}
\colorlet{pastelredmuted}{pastelred!75}
\colorlet{pastelpurplemuted}{pastelpurple!75}

\newenvironment{courier}{%
    \fontsize{7}{7}\fontfamily{pcr}\selectfont 
}{%
    \par 
}

\title{LLMs for Generating and Evaluating Counterfactuals: \\ A Comprehensive Study}

    \author{Van Bach Nguyen\textsuperscript{$\dag$}\Thanks{Equal contribution}$^*$  Paul Youssef\textsuperscript{$\dag$}$^*$   Christin Seifert\textsuperscript{$\dag$} Jörg Schlötterer\textsuperscript{$\dag\ddag$}\\
 \textsuperscript{$\dag$}University of Marburg, \textsuperscript{$\ddag$}University of Mannheim\\
  \texttt{\{vanbach.nguyen, paul.youssef, christin.seifert,} \\
  \texttt{joerg.schloetterer\}@uni-marburg.de}\\}

\begin{document}
\maketitle
\begin{abstract}
% As AI models become more complex, understanding their decisions is crucial. Counterfactual generation, where minimal changes to text flip a model's prediction, offers a way to explain these models. While Large Language Models (LLMs) have shown remarkable performance in NLP tasks, their efficacy in generating high-quality counterfactuals (CFs) remains uncertain.  This work  fills this gap by investigating how well LLMs generate CFs. We conduct a comprehensive comparison of several common LLMs, evaluating their performance on the counterfactual generation task, assessing both its intrinsic metrics for counterfactual explanation and its impact on downstream tasks.  Moreover, we analyze differences between human and LLM-generated CFs, providing insights for future research directions. Our study establishes a novel connection between intrinsic metrics and downstream task performance, offering a cost-effective evaluation method for NLP tasks and paving the way for more efficient research in this domain. Furthermore, we evaluate LLMs' ability to assess counterfactual explanations, aligning their evaluations with human judgments. This offers a novel, human-like evaluation method for LLMs.
As NLP models become more complex, understanding their decisions becomes more crucial. Counterfactuals (CFs), where minimal changes to inputs flip a model's prediction, offer a way to explain these models. While Large Language Models (LLMs) have shown remarkable performance in NLP tasks, their efficacy in generating high-quality CFs remains uncertain. This work fills this gap by investigating how well LLMs generate CFs for three tasks. We conduct a comprehensive comparison of several common LLMs, and evaluate their CFs, assessing both intrinsic metrics, and the impact of these CFs on data augmentation. Moreover, we analyze differences between human and LLM-generated CFs, providing insights for future research directions. Our results show that LLMs generate fluent CFs, but struggle to keep the induced changes minimal. Generating CFs for Sentiment Analysis (SA) is less challenging than NLI and Hate Speech (HS) where LLMs show weaknesses in generating CFs that flip the original label. This also reflects on the data augmentation performance, where we observe a large gap between augmenting with human and LLM CFs. Furthermore, we evaluate LLMs' ability to assess CFs in a mislabelled data setting, and show that they have a strong bias towards agreeing with the provided labels. GPT4 is more robust against this bias, but it shows strong preference to its own generations. Our analysis suggests that safety training is causing GPT4 to prefer its generations, since these generations do not contain harmful content. Our findings reveal several limitations and point to potential future work directions. % \TODO{HS findings}

\end{abstract}

\section{Introduction}
% Importance of CFs

% recent success of LLMs
The growing popularity of artificial intelligence (AI) and increasingly complex ``black-box'' models have triggered a critical need for interpretability. As \citet{miller_explanation_2019}  highlights, explanations often seek to understand why an event $P$ occurred instead of an alternative $Q$. Ideally, explanations should demonstrate how minimal changes to an instance could have led to different outcomes. In the context of textual data, this translates to introducing minimal modifications to the text through word additions, replacements, or deletions, to flip the label assigned by a given classifier. Counterfactual generation in NLP aims to foster an understanding of models, thereby facilitating their improvement~\cite{Kaushik-etal-2020-learning}, debugging~\cite{ross-etal-2021-explaining}, or rectification~\cite{balashankar-etal-2023-improving}.

% In the NLP field, Large Language Models (LLMs) have demonstrated remarkable performance across various tasks~\cite{maynez-etal-2023-benchmarking,srivastava2022beyond}. Despite advancements in counterfactual generation methods, the effectiveness of LLMs in generating counterfactuals (CFs) remains uncertain. This study addresses this gap by assessing the capability of LLMs to generate counterfactuals and identifying the most effective LLMs among different models. We compare several common LLMs of varying sizes and accessibility, evaluating their performance on the counterfactual generation task. We assess the quality of the generated counterfactuals based on standard metrics for counterfactual generation and evaluate the language quality of each LLM in the counterfactual generation task. Furthermore, we assess the generated counterfactuals on downstream tasks, such as data augmentation for sentiment analysis and natural language inference (SNLI) tasks.
\begin{figure}[t]
%trim:  left bottom right top, trim={5.4cm 21cm 0.5cm 4.3cm}
\includegraphics[width=\columnwidth]{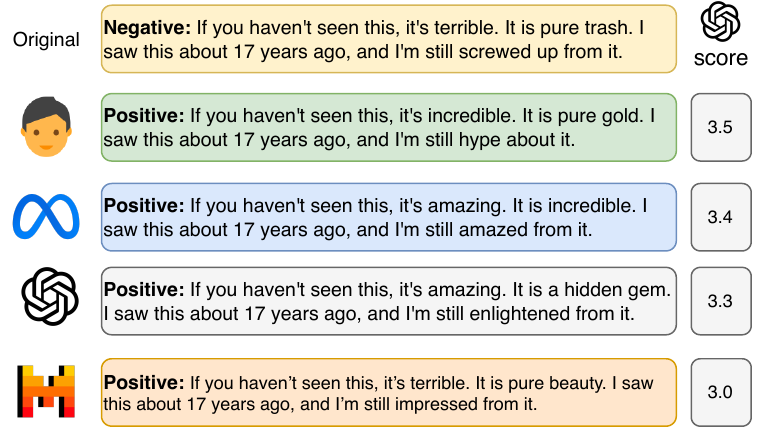}

%\caption{Counterfactual generation among different families  of Large Language Models ((LLAMA, Mistral, and GPT)) and Human Crowd. The corresponding models are LLAMA-2-7b-chat, GPT-3.5 and Mistral-56B. The scores represent the average of evaluations conducted with GPT-4.}
\caption{Counterfactual for Sentiment Analysis from several LLMs with their evaluation scores from GPT4.}
\label{fig:fig1}
\end{figure}
In the field of NLP, LLMs have consistently demonstrated remarkable performance across diverse tasks. However, despite significant advancements in counterfactual generation methods, the efficacy of LLMs in producing high-quality counterfactuals (CFs) remains an open question. Our study bridges this gap by rigorously assessing the inherent capability of LLMs to generate CFs and identifying the most effective ones. We conduct a comprehensive comparison of several common LLMs, spanning different sizes and accessibility levels, evaluating their performance specifically on the counterfactual generation task. Our assessment encompasses standard metrics for CFs quality, as well as an in-depth evaluation of language fluency tailored to the context of counterfactual generation. Furthermore, we extend our analysis to data augmentation. We consider generating CFs for 3 tasks in this study: Sentiment Analysis (SA), Natural Language Inference (NLI), and Hate Speech (HS).

Our analysis demonstrates that LLMs are able to generate fluent text. However, they have difficulties in inducing minimal changes. Generating CFs for SA is less challenging than NLI and HS, where LLMs exhibit weaknesses in generating CFs that flip the labels. For data augmentation, SA CFs from LLMs can be an alternative to human CFs, as they are able to achieve similar performance, while on NLI and HS further improvements are needed. Furthermore, we show a positive correlation between keeping minimal changes and data augmentation performance. This suggests a new direction to generate improved data for augmentation, potentially leading to more efficient augmentation approaches. % \TODO{HS findings}

We further assess the ability of LLMs to act as evaluators of CFs. We show a sample of CFs from different LLMs with the corresponding scores in Figure~\ref{fig:fig1}. By conducting controlled experiments, we show that LLMs have a strong bias to agree with the given labels, even if these labels are incorrect. GPT4 demonstrates strong preference to its own generations. Our analysis suggests that one reason for this preference is safety training, i.e., GPT4 prefers its own generations, because these generations do not contain any harmful content. Finally, to facilitate further research, we contribute a new dataset of CFs generated by various LLMs.\footnote{\url{https://github.com/aix-group/llms-for-cfs/}} %\TODO{GPT4 bias}

\section{Evaluation Methodology}
\label{sec:evaluation}
%\section{Evaluation Methodology}
We conduct a multi-faceted evaluation, considering several use cases where CFs could be beneficial.

\subsection{Intrinsic Evaluation} 

Given a fixed classifier \(f\) and a dataset with $N$ samples $(x_1, x_2, \ldots, x_N)$, $x = (z_1, z_2, \ldots, z_n)$ represents a sequence of \(n\) tokens with a ground truth label $y$. A valid counterfactual \( x' \) should: (1) achieve the desired target label $y'$ with (2) minimal changes, and (3) align with likely feature distributions~\cite{molnar2022}. 
To evaluate these three desiderata, we consider the intrinsic properties of \textit{Flip Rate}, \textit{Textual Similarity}, and \textit{Perplexity} as also suggested in a benchmark for counterfactual evaluation~\cite{nguyen-etal-2024-ceval-benchmark}: \\
%Therefore, in this evaluation, we consider the intrinsic properties of \textit{Flip Rate}, \textit{Textual Similarity}, and \textit{Perplexity} that correspond to each criterion, respectively: \\

\textit{Flip Rate (FR):} measures how effectively a method can change labels of instances with respect to a pretrained classifier. FR is defined as the percentage of generated instances where the labels are flipped over the total number of instances $N$~\cite{bhattacharjee2024llmguided}:
\[FR =\frac{1}{N} \sum_{i=1}^{N} \mathbbm{1}[f(x_i) = y'] \]\\
\textit{Textual Similarity (TS)}: quantifies the closeness between an original instance and the counterfactual. Lower distances indicate greater similarity.  We use the Levenshtein distance for \(d\) to quantify the token-distance between the original instance \(x\) and the counterfactual \(x'\). This choice is motivated by the Levenshtein distance's ability to capture all type of edits (insertions, deletions, or substitutions) and also its widespread use in related work~\cite{ross-etal-2021-explaining, treviso-etal-2023-crest}:
\[TS= \frac{1}{N} \sum_{i=1}^{N} \frac{{d(x_i, x'_i})}{|x_i|}\]\\
\textit{Perplexity (PPL):}
To ensure that the generated text is plausible, realistic, and follows a natural text distribution, we leverage perplexity from GPT-2 because of its effectiveness in capturing such distributions.~\cite{radford_language_nodate}\footnote{While GPT-2 is used for simplicity in this study, any other LLM can be substituted as long as it demonstrates strong text generation capabilities}
\[PPL(x) = \exp\left\{-\frac{1}{n} \sum_{i=1}^{n} \log p_{\theta}(z_i \mid z_{<i})\right\}\]
% Table 
% \textit{Diversity (Div):}
% We quantify diversity by measuring the token distance between pairwise generated instances. Given two counterfactuals, \(x'^1\) and \(x'^2\), for the same instance \(x\), diversity is defined as the mean pairwise distance between the sets of counterfactuals:
% \[ Div = \frac{1}{N} \sum_{i=1}^{N} \frac{d(x'^1_i, x'^2_i)}{|x_i|} \]
% Here, \(d(x'^1_i, x'^2_i)\) represents the Levenshtein distance between the corresponding tokens of the two counterfactuals for the \(i\)-th instance, and \(|x_i|\) is the length of the original instance \(x_i\).

\subsection{Data Augmentation}
After detecting failures in task-specific models, CFs can be used to augment the training data, and help close potential flaws in the reasoning of these models~\cite{Kaushik-etal-2020-learning}. Additionally, data augmentation with CFs increases generalization and OOD performance~\cite{sen-etal-2021-counterfactually, ding-etal-2024-data}. In this evaluation, we examine how augmenting original training data with human and LLMs-generated CFs reflects on the performance of task-specific models.

\subsection{LLMs for CFs Evaluation}
Evaluation with LLMs has been shown to be a valid alternative to human evaluation on various tasks like open-ended story generation and adversarial attacks~\cite{chiang-lee-2023-large}, open-ended questions~\cite{zheng-etal-2024-judging}, translation~\cite{kocmi-federmann-2023-large} and natural language generation~\cite{liusie-etal-2024-llm}. In this work, we examine how well LLMs can evaluate CFs. Detecting mistakes in CFs with LLMs opens the door for iteratively refining CFs~\cite{madaan-etal-2023-self-refine}. 

For assessing LLMs in CFs evaluation, we leverage them to evaluate two sets of CFs. An \emph{honest} set of CFs from humans, and a \emph{corrupted} set, where we corrupt the ground truth labels. We compare the scores between the two sets and draw conclusions about the realiablity of LLMs for evaluating CFs. %and show anecdotal evidence in  Section~\ref{subsec:llm_eval}.

\section{Experimental Setup}
\label{sec:experimental}

\subsection{Data}

We compare CFs generated by LLMs against CFs generated by crowd workers~\cite{Kaushik-etal-2020-learning} and experts~\cite{gardner-etal-2020-evaluating} (hereinafter referred to as ``Human Crowd'' and ``Human Experts'' respectively).

\paragraph{Sentiment Analysis (SA).} We experiment with the IMDb dataset~\cite{maas-etal-2011-learning}. For better comparability, we use the data splits from~\citet{Kaushik-etal-2020-learning}.

\paragraph{Natural Language Inference (NLI).} We experiment with SNLI~\cite{bowman-etal-2015-large}. Here too, we use the data splits from~\citet{Kaushik-etal-2020-learning}.

\paragraph{Hate Speech (HS).} We use the dataset from \cite{vidgen-etal-2021-learning}, which includes human CFs.

\subsection{Generating Countefactuals} %move to methodology
\label{subsec:cfs_generation}
In order to make our study LLMs-focused and computationally feasible, we decided to generate counterfactual in a way that fulfills the following criteria: 

% Criteria
% intendend for data augmentation
% does not require human intervention 
% no finetuning is requierd

\begin{itemize}
    \item Generated CFs can be used for data augmentation (an evaluation aspect)
    \item Generating CFs does not require human intervention (e.g., specifying edits or labeling)
    \item Generating CFs does not require additional training in order to make the study computationally feasible 
    \item The resulting CFs should depend only on the evaluated LLM in order to exclude any other confounding factors 
\end{itemize}

\begin{figure}[ht]
%trim:  left bottom right top, trim={5.4cm 21cm 0.5cm 4.3cm}
\includegraphics[width=\columnwidth, trim={1cm 4.5cm 2cm 1cm}, clip]{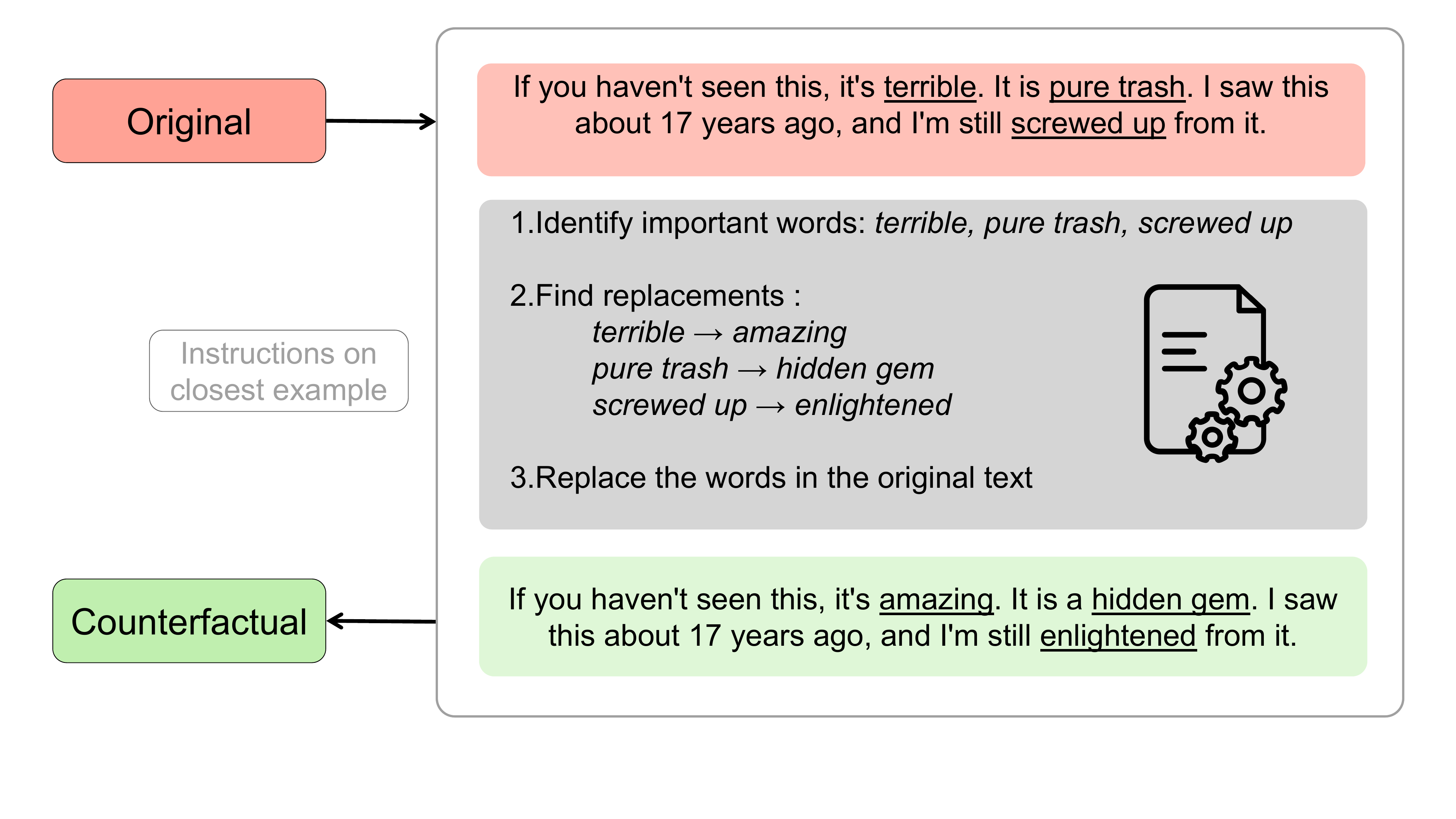}
\caption{An overview of CFs generation process. Step-by-step instructions are shown on closest example.}
\label{fig:cfs_generation}
\end{figure}

To create the prompt for the LLMs to generate CFs, we combine two techniques: (1) Selecting the closest factual instance to the current instance~\cite{liu-etal-2022-makes}. Since the provided example has a crucial effect on performance~\cite{liu-etal-2022-makes}, we select the closest factual/counterfactual pair that has been generated by humans. We use SentenceBERT~\cite{reimers-gurevych-2019-sentence} to obtain the latent space representation, and then calculate the distance using cosine similarity from that latent space. (2) Chain-of-Thought (CoT) prompting~\cite{wei-etal-2022-chain}, showing the necessary steps to generate a counterfactual instance based on a factual one, since it has been shown to help LLMs reason better and provide higher-quality answers. An overview of the process for generating CFs is depicted in Figure~\ref{fig:cfs_generation}.

Specifically, we use the validation set in each dataset as a reference to select the closest example when generating CFs for the train and test sets. After obtaining the pair of closest instances, we apply CoT prompting by defining three steps to generate the counterfactual: 
\begin{itemize}
    \item Step 1: Identify all of the important words that contribute to flipping the label.
    \item Step 2: Find replacements for the words identified in Step 1 that lead to the target label.
    \item Step 3: Replace the words from Step 2 in the original text to obtain the counterfactual instance. 
\end{itemize}
This prompt aligns with other work~\cite{ross-etal-2021-explaining, treviso-etal-2023-crest, li-etal-2024-prompting}, which involve identifying significant words that impact the label and altering them to flip the label, thereby generating counterfactual instances. The prompt examples can be found in the Appendix~\ref{app:prompt}.

\subsection{LLMs}
We compare open-source LLMs with closed-source LLMs. We choose LLAMA-2~\cite{touvron2023llama} and Mistral~\cite{jiang2023mistral} as representatives for open-source models, and GPT-3.5 and GPT-4\footnote{We use API from https://openai.com/} as representatives for closed-source LLMs. Table~\ref{tab:LLMs} summarizes the properties of each LLM.
\begin{table}[ht!]
\centering

\begin{tabular}{+l^l^c^c^c^c}
\toprule \tabhead
Model & Size &HF&Instruct& OS   \\\otoprule
LLAMA2& 7B/70B & \greencheck & \greencheck & \greencheck  \\
Mistral& 7B/56B &\redcross& \greencheck &\greencheck \\\midrule

GPT3.5& -&\greencheck& \greencheck & \redcross \\
GPT4& -&\greencheck&\greencheck &\redcross \\\bottomrule
\end{tabular}
\caption{Characteristics of Large Language Models (LLMs, including Size, Human Feedback (HF), Instruction, and Open-Source (OS). }
\label{tab:LLMs}
\end{table}
%\subsection{Experiment setup for robustness and augmentation}

\section{Results and Discussion}
\label{sec:results_discussion}

\subsection{Intrinsic Evaluation} 
\label{eval:intrinsic}
We show the results for the intrinsic evaluation in Table~\ref{tab:intrinsic}. For flip rate, we use SOTA BERT-based models from~\cite{morris-etal-2020-textattack} (SA and NLI) and~\cite{vidgen-etal-2021-learning} (HS). 

The obtained perplexity values reflect the high fluency of LLMs, some of which are even more fluent than humans.\footnote{Note that the shown perplexity values are based on GPT-2}The perplexity of HS is significantly higher than that of other datasets due to the informal nature of tweets, where users often use slang, uncommon words, or elongated words for emphasis. Distance values show that LLMs do not necessarily adhere to conducting minimal changes. One exception here is GPT3.5, whose average distance values resemble that of human-generated CFs. The large distance values for LLM-generated CFs could be explained by their tendency to overgenerate~\cite{guerreiro-etal-2023-hallucinations}.

In terms of flip rate, we notice that some LLM-generated CFs can have a higher flip rate than human-generated CFs on SA, whereas the opposite can be observed on NLI. Meanwhile, LLM-generated CFs can reach moderate FR in HS. NLI CFs could be more difficult to generate than SA and HS CFs, which explains the gap in flip rate between LLMs and humans on the one hand, and GPT4 and other LLMs on the other hand (this is especially apparent on the \emph{NLI - hypothesis}). This suggests that GPT4 should be the preferred choice to generate CFs for explaining a model's behavior. Furthermore, across all datasets, LLMs struggle to flip the label while keeping the changes minimal, i.e., they often need to make many modifications to flip the label. We examine the LLM-generated CFs in more detail in Section~\ref{subsec:qualitative}. %\TODO{HS findings, remove note about LLAMA2 70B, update table of success rate in appendix => Done}\st{Note that LLAMA2 70B appears to have a higher flip rate on the \emph{SA} and \emph{NLI - premise} test sets, but it also has a much lower success rate for generating CFs (cf. \mbox{Table~\ref{tab:robustness}} in \mbox{Appendix~\ref{app:robustness})}}.

%\st{We also notice that Mistral 56B, which has not be trained with instruction-tuning or RLHF, produces low quality CFs in terms of the metrics considered here. This highlights the importance of instruction-tuning and RLHF in generating high-quality texts. }\TODO{adapt/remove => Done}

\emph{This part of the evaluation shows us that LLMs are able to generate fluent CFs, but struggle to induce minimal changes. It also demonstrates that it is challenging to generate NLI and HS CFs that flip the label, whereas generating SA CFs is less difficult.} %\TODO{HS? => Done}

\begin{table*}[h]
\centering
\addtolength{\tabcolsep}{-1.5pt} 
\footnotesize
\begin{tabular}{lccccccccccccccc}
\toprule \tabhead
&\multicolumn{3}{c}{\textbf{SA}}&&\multicolumn{3}{c}{\textbf{NLI - premise}}&&\multicolumn{3}{c}{\textbf{NLI - hypothesis}}&&\multicolumn{3}{c}{\textbf{Hate Speech}}\\
\cmidrule{2-4}\cmidrule{6-8} \cmidrule{10-12} \cmidrule{14-16}
   & PPL $\downarrow$     & TS  $\downarrow$    & FR $\uparrow$     && PPL $\downarrow$     & TS  $\downarrow$    & FR $\uparrow$  &  & PPL $\downarrow$    & TS $\downarrow$     & FR $\uparrow$         && PPL $\downarrow$ & TS $\downarrow$ & FR $\uparrow$      \\
\otoprule
Human Experts   & 51.07 &  0.16 & 81.15 && - &- & - & &- &- & - && - &- & - \\

Human Crowd &  48.03 &  0.14 &  85.66 && 74.89 &  0.17 &     59.13 & &65.67 &  0.19 &     79.75 && 229.05 &  0.31 &   87.39\\
\midrule
GPT3.5& 49.53 &  \textbf{0.16} &     79.51 && 71.62 &  \textbf{0.15} &     35.50 & &51.30 &  \textbf{0.19} &     41.50  && 235.52 &  \textbf{0.16} &     54.05 \\

GPT4& 49.05 &  0.29 &     94.03 && 73.39 &  0.28 &     \textbf{57.12} & & 58.35 &  0.21 &     \textbf{65.88} && \textbf{209.49} &  0.49 &     \textbf{76.54} \\

LLAMA2 7B & 46.99 &  0.64 &     78.26 && 70.34 &  0.36 &     41.02& &59.60 &  0.28 &     38.64 && - &- &-\\
% \midrule
LLAMA2 70B & \textbf{33.88} &  1.37 &  93.48 && \textbf{63.17} &  0.21 &     41.07 && 58.54 &  0.23 &     46.62 && - &- & -  \\
Mistral 7B  & 48.55 &  1.06 &     95.13 && 78.34 &  0.36 &     37.71&& \textbf{39.06} &  0.46 &     44.11 && 365.15 &  0.69 &     67.41 \\
% &        Likability $\uparrow$ & 2.96  & 2.79 & 2.63 & 2.36 & 2.90 &2.84&& 3.19 & 2.77 & 1.90 & 2.35 & 3.01  \\
        Mistral 56B & 35.63 &  0.57 &     \textbf{95.45} && 65.37 &  0.23 &     27.46&& 57.65 &  0.21 &     31.55 && 401.63 &  0.56 &     70.30\\
\bottomrule
\end{tabular}
\addtolength{\tabcolsep}{1.5pt}
\caption{Metrics for intrinsic evaluation. \textbf{PPL} is perpelexity using GPT-2. \textbf{TS} is Levenshtein distance. \textbf{FR} is flip rate with respect to a SOTA classifier.\footnotemark}
\label{tab:intrinsic}
\end{table*}
\subsection{Data Augmentation}
\label{eval:augmentation}
We train on both original training data and CFs from different LLMs to see if augmenting the training data leads to an improved performance. For comparison, we conduct data augmentation with human CFs as well. The results for SA, NLI and HS are shown in Table~\ref{tab:sa:augmentation}, ~\ref{tab:nli:augmentation} and \ref{tab:speech:augmentation} respectively.

\paragraph{SA.} On the crowd CFs and expert CFs test sets for SA including LLM-generated CFs lead to improved performance. LLAMA2 7B provide the most useful CFs for data augmentation, but other LLMs perform similarly. However, augmenting with human CFs works the best. On the original test set, augmenting with CFs does not improve performance. This shows that the gains in performance from data augmentation are visible only if the test set contains challenging examples.

\iffalse
\begin{table}[ht]
    \centering       %\footnotesize
\begin{tabular}{lccc}
\toprule
&\multicolumn{3}{c}{\textbf{Test Data}}\\
\cmidrule{2-4}
  & \makecell{\textbf{crowd} \\ \textbf{CFs}} & \makecell{\textbf{expert} \\ \textbf{CFs}} & \textbf{original} \\
\midrule
Original only & 92.21 & 86.27 & \textbf{90.16} \\
Human Crowd & \textbf{95.90} & \textbf{92.62} & 89.96 \\
\midrule
GPT3.5 & 95.08 & \textbf{91.19} & 89.96 \\
GPT4 & 93.65 & 88.73 & \textbf{90.37} \\
LLAMA2 70B & 94.26 & 90.78 & 89.55 \\
LLAMA2 7B & \textbf{95.49} & \textbf{91.19} & 89.75 \\
Mistral 7B & 92.42 & 88.11 & 88.93 \\
Mistral 56B & 93.85 & 86.89 & 84.02 \\

\bottomrule
\end{tabular}
\caption{Data augmentation results for SA. Classification model is trained on original and LLMs or human-generated CFs. Evaluation is done on 3 test sets: crowd CFs, expert CFs, original with accuracy as a metric.}
    \label{tab:sent:augmentation:old}
\end{table}
\fi
% This shows that augmenting with CFs does not necessary lead to an improved performance.

 \paragraph{NLI.} On the \emph{crowd premise} test set of NLI, which consists of CFs that were generated by changing the premise only, we notice that most of the LLM-generated CFs help improve the model's performance by a good margin (> 7 pp). The gap to augmenting with human CFs, however, is still large ($\sim 9$ pp). On the \emph{crowd hyothesis} test set, all LLMs lead to a lower performance. Here too, there is a large gap to human CFs ($\sim16$ pp). On the \emph{original} test set, augmenting with LLM-generated CFs hurts performance, while augmenting with human-generated CFs bring good improvements ($\sim 5$ pp). This shows how high-quality human CFs improve the model's capabilities, and points to a problem with LLM-generated CFs for NLI.

\paragraph{HS.} Training with LLM-generated CFs does not bring substantial improvements on the CFs and the original test sets. Conversely, training with human CFs leads to significant improvements on both test sets. On this task too, LLM-generated CFs fall short of human CFs, indicating that there remains significant room for improvement.

\iffalse
\begin{table}[ht]
    \centering         
    \addtolength{\tabcolsep}{-3.5pt} 
\begin{tabular}{lccc}

\toprule
&\multicolumn{3}{c}{\textbf{Test Data}}\\
\cmidrule{2-4}
& \makecell{\textbf{crowd} \\ \textbf{Premise}} & \makecell{\textbf{crowd} \\ \textbf{Hypothesis}} & \textbf{original} \\
\midrule
Original only & 39.88 & 58.50 & 73.50 \\ 
Human Crowd   & \textbf{65.38} & \textbf{71.25} & \textbf{78.50} \\\midrule
GPT3.5   & 55.62 & 51.88 & 55.00 \\
GPT4   & 55.88 & \textbf{58.63} & \textbf{70.75} \\
LLAMA2 7B  & \textbf{56.25} & 51.12 & 57.50 \\
LLAMA2 70B  & 53.50 & 58.13 & 65.75 \\
Mistral 7B   & 52.88 & 48.38 & 57.75 \\
Mistral 56B   & 43.75 & 39.75 & 45.75 \\
\bottomrule

\end{tabular}
\addtolength{\tabcolsep}{3.5pt} 
    \caption{Data augmentation results for NLI. Classification model is trained on original and LLMs or human-generated CFs with accuracy as a metric.}
    \label{tab:nli:augmentation:old}
\end{table}
\fi

% New results 

\footnotetext{LLAMA-2 is unable to generate counterfactuals for HS due to its safety mechanism.}
\paragraph{Connection with intrinsic metrics. } We examine the relation between data augmentation performance on the one hand and perplexity and Levenshtein distance on the other hand. The correlation values in Table~\ref{tab:corr:int_aug} suggest that CFs with lower distance (to the factual instances) bring more improvements. Indeed, classifiers could be insensitive to small changes~\cite{glockner-etal-2018-breaking}, and having such examples in the training can make classifiers more robust. 
The negative correlation between accuracy and perplexity suggests that more fluent CFs are less effective in improving the classifier's performance. This indicates that classifiers primarily focus on the content rather than grammatical structure or coherence, especially in NLI tasks where the (factual) instances are mere image captions that are not necessarily fluent or grammatical texts

\emph{In summary, most LLMs produce CFs that come close to human CFs in terms of data augmentation performance on SA. On NLI and HS, the results are less positive: LLM-generated CFs bring no improvements in most cases, and the gap to human CFs is still large. CFs with less changes to the factual instances are more beneficial for data augmentation.}

\begin{table}[ht]
    \centering 
    \small
    \addtolength{\tabcolsep}{-3.5pt} 
        \begin{tabular}{lccc}
        
        \toprule
        &\multicolumn{3}{c}{\textbf{Test Data}}\\
        \cmidrule{2-4}
        & \makecell{\textbf{Crowd} \\ \textbf{CFs}} & \makecell{\textbf{Expert} \\ \textbf{CFs}} & \textbf{Orig.} \\
        \midrule
            Original only & 91.68 ± 1.07 & 86.31 ± 1.62 & \textbf{90.20} ± 0.67 \\
              Human Crowd & \textbf{95.94} ± 0.37 & \textbf{92.01} ± 1.09 & 89.63 ± 0.85 \\ \midrule
                   GPT3.5 & 94.55 ± 0.96 & 89.88 ± 1.47 & 89.30 ± 0.51 \\
                    GPT4 & 93.52 ± 0.89 & 89.10 ± 0.76 & \textbf{89.88} ± 0.57 \\
    
                LLAMA2 7B & \textbf{95.29} ± 0.72 & \textbf{90.37} ± 1.57 & 88.89 ± 1.35 \\
               LLAMA2 70B & 94.18 ± 0.27 & 88.89 ± 1.02 & 89.39 ± 0.44 \\
               Mistral 7B & 93.93 ± 0.62 & 88.61 ± 1.68 & 89.22 ± 0.72 \\
              Mistral 56B & 93.40 ± 1.02 & 88.20 ± 0.79 & 89.84 ± 0.79 \\
        \bottomrule
        
        \end{tabular}
\addtolength{\tabcolsep}{3.5pt} 
    \caption{Data augmentation results for SA. Classification model is trained on original and LLMs or human-generated CFs with Accuracy as a metric.}
    \label{tab:sa:augmentation}
\end{table}

\begin{table}[ht]
    \centering 
    \small
    \addtolength{\tabcolsep}{-3.5pt} 
\begin{tabular}{lccc}

\toprule
&\multicolumn{3}{c}{\textbf{Test Data}}\\
\cmidrule{2-4}
& \makecell{\textbf{crowd} \\ \textbf{Premise}} & \makecell{\textbf{crowd} \\ \textbf{Hypothesis}} & \textbf{Orig.} \\
\midrule
           Original only &  43.60 ± 3.87 &     59.75 ± 3.06 & 71.85 ± 1.33 \\
             Human Crowd &  \textbf{63.42} ± 2.74 &     \textbf{70.53} ± 1.02 & \textbf{76.65} ± 2.04 \\ \midrule
               GPT3.5 &  \textbf{}54.42 ± 1.86 &     49.68 ± 2.64 & 53.00 ± 2.61 \\
                   GPT4 &  53.10 ± 1.85 &     \textbf{54.50} ± 1.28 & \textbf{63.50} ± 1.31 \\
              LLAMA2 7B &  52.85 ± 1.29 &     49.45 ± 2.03 & 58.15 ± 2.23 \\
             LLAMA2 70B &  \textbf{54.58} ± 3.69 &     49.02 ± 2.96 & 58.05 ± 0.78 \\
             Mistral 7B &  51.05 ± 2.89 &     46.52 ± 2.51 & 58.50 ± 2.50 \\
            Mistral 56B &  51.35 ± 1.79 &     45.45 ± 1.07 & 48.65 ± 1.88 \\
%Mixtral 56B\footnote{non-instruct}&  44.42 ± 3.55 &     45.92 ± 4.62 & 55.15 ± 2.57 \\
\bottomrule

\end{tabular}
\addtolength{\tabcolsep}{3.5pt} 
    \caption{Data augmentation results for NLI. Classification model is trained on original and LLMs or human-generated CFs with Accuracy metric.}
    \label{tab:nli:augmentation}
\end{table}\textbf{}
\begin{table}[ht]
    \centering 
        \small

    \addtolength{\tabcolsep}{-3.5pt} 
\begin{tabular}{p{2.5cm}P{2.25cm}P{2.25cm}}
\toprule
&\multicolumn{2}{c}{\textbf{Test Data}}\\
        \cmidrule{2-3}
   &          \textbf{CFs} &     \textbf{Orig.} \\ \midrule
Original only & 83.27 ± 2.66 & 70.28 ± 0.60 \\
        Human & \textbf{94.27} ± 0.20 & \textbf{94.30} ± 0.14 \\ \midrule
       GPT3.5 & 81.00 ± 2.87 &\textbf{ 70.29} ± 0.96 \\
         GPT4 & \textbf{86.00} ± 3.20 & 69.33 ± 0.49 \\
   Mistral 7B & 84.32 ± 2.52 & 69.90 ± 0.75 \\
  Mistral 56B & 82.86 ± 1.78 & 68.58 ± 0.97 \\
\bottomrule
\end{tabular}

\addtolength{\tabcolsep}{3.5pt} 
    \caption{Data augmentation results for Hate Speech. Classification model is trained on original and LLMs or human-generated CFs with accuracy as a metric.}
    \label{tab:speech:augmentation}
\end{table}
\begin{table}[h]
    \centering       
    % \footnotesize
        \begin{tabular}{p{3cm}p{1cm}p{1cm}p{1cm} }
        \toprule
                     \textbf{Compared values} &  \textbf{SA} &  \textbf{NLI}   &  \textbf{HS}\\
        \midrule

             Accuracy \& $-$PPL &  -0.26 &     -0.56 &  -0.10\\
            Accuracy \& $-$TS   &   0.49 &      0.52 &   0.60\\
        \bottomrule
        \end{tabular}
    \caption{Spearman correlations between intrinsic metrics and data augmentation performance.}
    \label{tab:corr:int_aug}
\end{table}

\subsection{LLMs for CFs Evaluation}\label{subsec:llm_eval}
We examine how reliable are LLMs for CFs evaluation by asking them to evaluate two sets of human CFs: an \emph{honest} set and a \emph{corrupted} set. The ``honest set'' refers to a collection of human CFs, for which the ground truth labels are provided, whereas the ``corrupted set'' consists of instances, for which wrong labels differing from the gold labels are provided. In the context of NLI, the third label, distinct from both the target and factual labels, is selected for inclusion in the corrupted set. For SA, the reverse label is chosen while the factual label remains undisclosed. Initially, we prompt GPT3.5 and GPT4 to assess whether the provided CFs accurately represent the target labels by assigning a score from 1 to 4 (cf. Appendix~\ref{app:prompt}). Here, a score of 1 or 2 indicates disagreement (complete or partial) with the target label, while a score of 3 or 4 indicates agreement (partial or complete) with the target label. Ideally, the evaluation LLMs should give high scores to the honest set, and low scores to the corrupted set. We show the distributions for disagreements and agreements in Table~\ref{tab:llm_eval:fl}. 

On SA, both LLMs perform well, but GPT4 exhibits higher sensitivity to the corrupted examples. On NLI, we notice that GPT3.5 gives high flip label scores to humans CFs with both correct and incorrect labels. GPT4 performs much better, but still exhibits high tendency to agree with wrong labels ($\sim$ 40\%). The results can be explained by the tendency of LLMs to agree with the provided answers, especially on reasoning tasks~\cite{zheng-etal-2024-judging}. To verify this, we prompt both LLMs to classify the same set of NLI CFs by choosing one of the three labels (entailment, neutral, contradiction) using a similar prompt. The classification results in Table~\ref{tab:llm_eval:cls} show an improved performance compared to asking the same LLMs if they agree with incorrect labels (cf. Table~\ref{tab:llm_eval:fl}). We also compare the flip label score distributions of GPT3.5 and GPT4 on the corrupted set in Table~\ref{tab:corrputed:dist}, and observe that even though GPT3.5 gives high scores to corrupted inputs it is less certain (most frequent score is 3), whereas GPT4 tends to be more certain and assigns mostly 1 or 4 (> 93\%). % Conclusion: Different prompt might help? ?

%We further compare the annotations of GPT4 to human annotations on a randomly selected sample. More specifically, we select 4 NLI CFs for each LLM, we also include 4 examples from human CFs. This gives us a sample of 28 examples. Two of the authors annotated this sample separately, and then resolved any inconsistencies. Comparing the agreed-upon annotations with 

%We further compare the annotations of GPT4 to human annotations on a randomly selected sample. More specifically, we select 4 NLI CFs for each LLM, we also include 4 examples from human CFs. This gives us a sample of 28 examples. Two of the authors annotated this sample separately using the same instructions provided to GPT4, and then resolved any inconsistencies. The spearman correlation between the agreed-upon annotations and the annotations from GPT4 is 0.70. 

\paragraph{Evaluation with GPT4.} We conduct a wide-scale CFs evaluation with GPT4. Besides verifying the target label \textbf{FL}, we also ask GPT4 to judge if there are any unnecessary alterations \textbf{UA}, and if the CF is realistic \textbf{RS}. For these aspects, we use a scoring scheme ranging from 1 to 4, where higher scores indicate better performance. The results for the GPT4 evaluation can be found in Table~\ref{tab:GPT4_eval}.

% Comment on table 
The evaluation scores from GPT4 show that GPT4 prefers LLM-CFs, and especially its own generations, which are given the highest scores on most datasets. On SA, Mistral 56B scores the highest with LLAMA 70B and GPT4 having slightly lower scores. On NLI, human CFs take the second position after GPT4. On HS, GPT4 performs the best, while human CFs are given the second lowest score on average. GPT4 might have a bias to prefer its own generations~\cite{panickssery-etal-2024-llm}. We further investigate this bias in Section~\ref{subsubsec:gpt_bias}. To further verify the evaluation scores from GPT4, we calculate the correlations between GPT4 scores and the scores from the intrinsic evaluation. 

% \st{The scores for SA are generally lower, which might be explained by the fact that SA is an easier task than NLI, and therefore evaluating SA CFs is also easier for GPT4, and potentially more accurate.} \TODO{mention HS => Done}

%\st{The correlations shown in \mbox{Table~\ref{tab:corr:int_GPT4}} reflect strong relation on SA with respect to all three aspects (flipping label, minimal changes, realisticness). On NLI, we notice the same pattern except for realisticness/perplexity, where we notice a weak negative relation. This might be due to the nature of NLI instances that are short image captions, and are not necessarily grammatical. } 
The correlations in Table~\ref{tab:corr:int_GPT4} indicate strong correlation for label flipping on SA and NLI, but weak correlation on HS. This suggests that GPT4 highly agrees with the classifier. Minimal changes show weak correlation with Levenshtein distance on SA and HS, with moderate correlation on NLI, implying that GPT4 is not necessarily sensitive to small changes. GPT4 shows weak to moderate positive correlation on realisticness with perplexity on HS and SA, and moderate negative correlation on NLI. This discrepancy might be due to the nature of the different texts, i.e., while SA contains long movie reviews, NLI contains short image captions and HS contains tweets.

%, suggesting text length may influence results (SA longest, HS intermediate, NLI shortest).}

%This clearly shows that GPT3.5 has a strong bias towards giving high scores, and is therefore unsuitable for evaluating if the provided CF indeed has a certain label. 

\emph{LLMs show a high tendency to agree with the provided labels even if these are incorrect, especially on tasks that require reasoning such as NLI.  The correlation between GPT4 evaluation scores and automated metrics for label flipping, textual distance, and fluency varies across tasks .} 
% \st{GPT4 evaluation scores correlate with automatic metrics for flipping label, distance and fluency.}
%     \label{tab:corr:int_GPT4}

\begin{table}[]
    \centering        
\begin{tabular}{llrrr}
\toprule
      \textbf{LLM/Set} &   \textbf{Task}  &  \textbf{1\&2} &   \textbf{3\&4} & \textbf{Avg.} \\
\midrule 

    \textbf{GPT3.5} &  &  &  & \\
\midrule

      Honest & SA & 3.61 & 96.39 &  3.43 \\
Corrupted & SA & 77.42 & 22.58 &  1.61 \\ \midrule

      Honest &    premise &  0.63 & 99.37 &  3.57 \\
Corrupted &    premise &  5.56 & 94.44 &  3.13 \\ \midrule
      Honest & hypothesis &  1.38 & 98.62 &  3.56 \\
Corrupted & hypothesis &  3.53 & 96.47 &   3.28 \\\midrule
 \midrule

      \textbf{GPT4} &  &  &  & \\
\midrule
      Honest &  SA & 7.53 & 92.47 &  3.66 \\
Corrupted & SA & 98.93 &  1.08 &  1.04 \\\midrule

       Honest &    premise & 12.31 & 87.69 &  3.58 \\
 Corrupted &    premise & 59.51 & 40.50 &  2.19 \\ \midrule
       Honest & hypothesis &  4.50 & 95.50 &  3.81 \\
 Corrupted & hypothesis & 57.87 & 42.12 &  2.29 \\

\bottomrule
\end{tabular}
    \caption{Flip label scores distribution for GPT3.5 and GPT4 on honest and corrupted sets.}
    \label{tab:llm_eval:fl}
\end{table}

% Move to appendix?
\begin{table}[ht]
    \centering         
    % \footnotesize
    % \small
        \begin{tabular}{p{1cm}p{2cm}p{2cm}p{1cm}}
        \toprule
          \textbf{Set} &    \textbf{LLM}   &   \textbf{Part} &   \textbf{Acc.} \\
        \midrule
        Honest & GPT3.5 & premise & 54.90 \\
            Honest & GPT3.5& hypothesis & 63.08 \\ \midrule 

            Honest & GPT4  & premise & 59.25 \\  
            Honest & GPT4 & hypothesis & 75.75 \\

        \bottomrule
        \end{tabular}
    \caption{Classification performance on human CFs. Note the improved performance compared to asking LLMs if they agree with a given label (cf. Table~\ref{tab:llm_eval:fl}).}
    \label{tab:llm_eval:cls}
\end{table}

\begin{table*}[h]
\centering
\addtolength{\tabcolsep}{-2.7pt} 
\footnotesize
\begin{tabular}{lccccccccccccccccccc}
\toprule \tabhead
&\multicolumn{4}{c}{\textbf{SA}}&&\multicolumn{4}{c}{\textbf{NLI - premise}}&&\multicolumn{4}{c}{\textbf{NLI - hypothesis}}&&\multicolumn{4}{c}{\textbf{Hate Speech}}\\
\cmidrule{2-5}\cmidrule{7-10} \cmidrule{12-15} \cmidrule{17-20}
   & FL     & UA    & RS & Avg.     && FL     & UA    & RS & Avg.  &  & FL     & UA    & RS & Avg. && FL     & UA    & RS & Avg.      \\
\otoprule
Expert Humans   & 3.54 &                     2.69 &                 2.49 & 2.91 && - & - &- & - & &- & - &- & - && - & - &- & - \\

Crowd Humans &  3.66 &                  2.95 &                 2.58 & 3.06 && \underline{3.58} &                   \textbf{3.88} &              \textbf{3.86}  & \underline{3.77} & &\underline{3.81} &      \underline{3.96} &        3.81 & \underline{3.86} && 3.04 &                   3.54 &              3.19 & 3.26\\
\midrule
GPT3.5&  3.58 &                     2.91 &                 2.65 & 3.05 && 2.51 &                     3.82 &                 3.69           & 3.34 & &3.19 &                     3.93 &                 3.74 & 3.62 && 1.78 &                   \underline{3.58} &              3.02 & 2.79 \\

GPT4& 3.79 &            \textbf{3.15} &     2.91 & 3.28 && \textbf{3.68} &             \underline{3.83} &           \underline{3.84}          & \textbf{3.78} & &\textbf{3.96} &             \textbf{3.98} &       \textbf{3.92} &  \textbf{3.95} && \textbf{3.65} &                   \textbf{3.73} &              \textbf{3.63} & \textbf{3.67} \\

LLAMA2 7B & 3.60 &                     2.74 &                 2.63 & 2.99 && 2.96 &                     3.38 &                 3.67           & 3.34 & &3.23 &                     3.74 &                 3.66 & 3.54 && - & - &- & -\\
% \midrule
%LLAMA2 70B & 3.70 &                    2.75 &                 2.47 & 2.97 && 3.35 &                     3.46 &                 3.67          & 3.49 & &3.49 &                     3.68 &                 3.56 & 3.58&& - & - &- & -  \\
LLAMA2 70B &  \underline{3.87} &                     3.05 &                 \textbf{2.96} & \underline{3.29} &&3.07 &                     3.68 &                 3.77 & 3.51 & &3.60 &                     3.89 &                 3.75 & 3.75 && - & - &- & -  \\
Mistral 7B  &  3.85 &                    2.84 &        2.69 & 3.13 && 2.97 &                     3.63 &                 3.74           & 3.45 & &3.50 &                     3.70 &                 3.65 & 3.62 && \underline{3.32} &                   \underline{3.58} &              \underline{3.40} & \underline{3.43} \\
%Mistral 56B & 2.58 &                    1.74 &                 1.75 & 2.02 && 2.37 &                     3.11 &                 3.49          & 2.99 & &3.02 &                     3.45 &                 3.48 & 3.32 && 3.31 &                   3.44 &              3.25 & 3.33\\
Mistral 56B & \textbf{3.88} &                     \underline{3.07} &                 \underline{2.94} & \textbf{3.30} && 2.71 &                     3.81 &                 3.75 & 3.42 & &2.95 &                     3.94 &                 \underline{3.84}     & 3.58 && 3.31 &                   3.44 &              3.25 & 3.33\\
\bottomrule
\end{tabular}
\addtolength{\tabcolsep}{2.7pt}
\caption{Scores for evaluation with GPT4. \textbf{FL} refers to flipping label score, \textbf{UA} to unncessary alteration, \textbf{RS} is the realisticness score, and \textbf{Avg.} is the average of the three scores. Best score for each task is in \textbf{bold}. Second best score is \underline{underlined}.}
\label{tab:GPT4_eval}
\end{table*}

\iffalse

\begin{table}[ht]
    \centering         

        % \footnotesize

\begin{tabular}{p{3cm}p{2cm}r}
\toprule
            \textbf{Compared Values} &  \textbf{SA} &  \textbf{NLI} \\
\midrule
\textbf{FL \& FR} &  0.83 &      0.92 \\
\textbf{UA \& -TS} &  0.50 &      0.75 \\
\textbf{RS \& -PPL} &  0.62 &     -0.23 \\
\bottomrule
\end{tabular}
    \caption{Spearman correlations between intrinsic metrics and GPT4 evaluation scores. \textbf{PPL} and \textbf{TS} scores are negated so that higher is better.}
    \label{tab:corr:int_GPT4}
\end{table}
\fi

\begin{table}[ht]
    \centering         

        % \footnotesize

    \begin{tabular}{p{3cm}p{1cm}p{1cm}p{1cm}}
    \toprule
    \textbf{Compared Values} &  \textbf{SA} &  \textbf{NLI}  & \textbf{HS} \\
    \midrule
    \textbf{FL \& FR}  &  0.86 &      0.92 &   0.30\\
    \textbf{UA \& -TS}  & 0.18 &      0.60 &   0.10\\
    \textbf{RS \& -PPL} & 0.43 &     -0.26 &   0.20\\
    \bottomrule
    \end{tabular}
    \caption{Spearman correlations between intrinsic metrics and GPT-4 evaluation scores. \textbf{PPL} and \textbf{TS} scores are negated so that higher is better.}
    \label{tab:corr:int_GPT4}
\end{table}

\iffalse
\begin{table}[ht]
    %\footnotesize

    \centering         
    % \footnotesize
        \begin{tabular}{rrr}
        \toprule
         \textbf{Score} &  \textbf{GPT3.5} &  \textbf{GPT4} \\
        \midrule
        1 &    0.70 & 55.50 \\
         2 &    3.85 &  3.19 \\
         3 &   69.61 &  2.94 \\
         4 &   25.84 & 38.37 \\
        \bottomrule
        \end{tabular}
    \caption{Flip label score distributions for GPT3.5 and GPT4 on the corrupted set NLI. Distribution is an average of the distributions on the premise and hypothesis sets.}
    \label{tab:corrputed:dist}
\end{table}
\fi

\begin{table}[]
    \centering        
        % \footnotesize 
        \begin{tabular}{p{1.5cm}P{1cm}P{1cm}P{1cm}P{1cm}}
        \toprule
         \textbf{LLM/Score} &  \textbf{1}  & \textbf{2} & \textbf{3} & \textbf{4}  \\
        \midrule
        \textbf{GPT3.5} & 0.70 &    3.85  & 69.61  &  25.84 \\
         \textbf{GPT4} & 55.50 &    3.19  &  2.94  & 38.37 \\
         %3 &    &   \\
         %4 &    &  \\
        \bottomrule
        \end{tabular}
    \caption{Flip label score distributions on the corrupted set of NLI. Distribution is an average of the distributions on the premise and hypothesis sets.}
    \label{tab:corrputed:dist}
\end{table}

\subsection{Qualitative Analysis}
\label{subsec:qualitative}
\subsubsection{CFs for NLI}
We look into a selected set of examples based on the evaluation from GPT4. For each LLM, we pick 2 NLI examples with the highest/lowest scores. We end up with 28 examples. We identify three categories of errors based on this sample : 
\begin{itemize}
    \item \textbf{Copy-Paste:} When asked to generate a CF, and change the label from \emph{contradiction to entailment}, LLMs will use the unchanged part (premise or hypothesis) as output. This a clever but lazy way to flip the label to \emph{entailment}, since two identical sentences would naturally have the label \emph{entailment}. These CFs were given perfect scores by GPT4. Table~\ref{tab:copypaste} in the Appendix shows the percentage of copy-paste CFs for all LLMs (at most 4.27\% for GPT3.5).
    
    \item \textbf{Negation:} When asked to to change the label from \emph{entailment to contradiction}, LLMs would negate the premise/hypothesis. The negation does not make sense in the observed CFs.%, and GPT4 rightly assigns them the lowest scores possible. 

    \item \textbf{Inconsistency:} These examples contain contradictory or illogical sentences, but GPT4 sometimes incorrectly assigned high scores. 
    
\end{itemize}

We show examples for each error category in Table~\ref{tab:categorized_examples}. % TODO 

\subsubsection{Evaluation Scores}
We also look into the evaluation scores from GPT4 on the same set of examples. We show correct and incorrect evaluations in Table~\ref{tab:eval_examples}. GPT4 assigns high scores to contradictory examples, which partially fulfill the target label, and low scores to examples which contain valid minimal changes. GPT4 could be insensitive to such small changes. 

\subsubsection{Bias in GPT4 Scores}
\label{subsubsec:gpt_bias}
Given GPT4's preference towards its own generations (cf. Table~\ref{tab:GPT4_eval}), we conduct a qualitative analysis to examine if we agree with the scores given by GPT4 on a set of SA CFs. More specifically, we examine a set of expert CFs that were given lower scores than their corresponding GPT4 CFs on all three metrics. On 12 out of 14 instances we do not agree with the scores given by GPT4. We notice that GPT4 unnecessarily changes some parts of the movie reviews, and introduces changes that do not make sense in the wider context of the reviews. We also noticed that GPT4 changes/omits parts containing potentially harmful content (e.g., torture, sexual content, etc.). Hence, we believe GPT4 prefers its own generations, because these generations do not contain any harmful content (despite safety not being an evaluation criteria).

\begin{table*}[]
    \centering
    \tiny
\begin{tabular}{p{2cm}p{2cm}p{2cm}llllllll}
\toprule
\textbf{Original Premise} &    \textbf{Original Hypothesis} & \textbf{ Counterfactual} & \textbf{Original Label} & \textbf{Target Label} & \textbf{FL} & \textbf{UA} & \textbf{RS} &         \textbf{LLM} &               \textbf{Changed Part} \\
\midrule
\textbf{Copy-Paste} &                                                                    &                                                                                                                         &                &                &            &                  &              &             &                    \\ \midrule
A photographer and assistant take a photo of a man in a suit and a woman in a dress in front of large white columns and a yellow wall. &                       The photographer is taking a photo of a dog. &                                                                            The photographer is taking a photo of a dog. &  contradiction &     entailment &        4.0 &              4.0 &          4.0 &        GPT4 &    premise \\
                                                   At least eight people are sitting at a table in a building that overlooks the city. &                                        People sitting on the floor &                        At least eight people are sitting at a table on the floor in a building that overlooks the city. &  contradiction &     entailment &        4.0 &              4.0 &          4.0 &     Mistral 7B & hypothesis \\ \midrule
                                                                                                                      \textbf{Inconsistency} &                                                                    &                                                                                                                         &                &                &            &                  &              &             &                    \\ \midrule
                                                   At least eight people are sitting at a table in a building that overlooks the city. &                                        People sitting on the floor & At least eight people are sitting at a table in a building that overlooks the city. People can be sitting on the floor. &  contradiction &     entailment &        4.0 &              4.0 &          4.0 &     Mistral 7B &    premise \\
                                                                   An old man in a blue shirt and white pants glowers by a glass door. &                                              A man resting outside &                                                                                                  A glass man by a door. &        neutral &  contradiction &        1.0 &              1.0 &          1.0 &       Human Crowd & hypothesis \\ 
% \iffalse
\midrule
\textbf{Negation} &    &    &    &   &   &    &   &    &                    \\ \midrule
%A young man standing outside a laundromat. &                                     A man is standing. &                      A middle-aged man is standing in a restaurant. &     entailment &  contradiction &        1.0 &              1.0 &          4.0 & Mistral 56B &    premise \\
                               Two men in costumes with fake carrot noses, top hats, sunglasses and white fur coats that contain white electrical lights. &   People in costumes &                   Two women in costumes with real carrot noses, no hats, no sunglasses, no coats, no lights. &     entailment &  contradiction &        1.0 &              2.0 &          4.0 &  LLAMA2 70B &    premise \\
                          % A man with a beard is talking on the cellphone and standing next to someone who is lying down on the street. & A man is prone on the street while another man stands next to him. &                                                                  Unproning on the street while another man lies in bed. &     entailment &  contradiction &        1.0 &              1.0 &          1.0 & mistral\_56b & hypothesis \\
% \fi
\bottomrule
\end{tabular}
    \caption{Categorization of a sample of incorrect NLI CFs with evaluation scores from GPT4. }
    \label{tab:categorized_examples}
\end{table*}

%\section{Further Analysis}
%\label{sec:analysis}
%\input{src/analysis}

\section{Related Work}
\label{sec:related_work}
% 1. Methods for counterfactuals generation
% \cite{ross-etal-2021-explaining} % MICE
% \cite{robeer-etal-2021-generating-realistic} 
% \cite{wu-etal-2021-polyjuice} % polyjuice
% \cite{chen-etal-2023-disco} % DISCO
% \cite{treviso-etal-2023-crest} % CREST

\paragraph{Large Language Models.}
%  Ability of LLMs has been evaluated {cite} but not cf counterfactual, 
% difference in performance between closed-source and open-source LLms => compare both
% number of
LLMs have demonstrated impressive capabilities across a diverse natural language processing tasks, such as question answering, wherein the model needs to retrieve relevant information from its training data and generate a concise response, or text summarization, which distills lengthy texts into concise summaries while retaining crucial information~\cite{maynez-etal-2023-benchmarking}. However, the task of CFs generation has not been comprehensively evaluated for LLMs.  A large number of LLMs exist, exhibiting variations in model size, architecture, training dataset, the incorporation of human feedback loops and accessibility (open-source or proprietary)~\cite{zhao2023survey}. Consequently, there is a necessity to conduct comparative evaluations across different models on a standardized task. Since the architectures of the LLMs under consideration are predominantly similar, and the training datasets are either known public sources or undisclosed, the primary focus of this study is to compare LLMs that are different in model size, the implementation of human feedback, and accessibility. To enhance the performance of LLMs across various tasks, in-context learning (ICL) techniques have been employed to optimize the prompts provided to these models. Numerous prompt engineering approaches during the inference phase have been proposed, either by selecting the demonstration instances, or formatting the prompt in form of instruction or reasoning steps ~\cite{dong2022survey}. In this study, leverage chain-of-thought prompting (CoT)~\cite{wei-etal-2022-chain} and selecting closest instance retrieval strategies\cite{liu-etal-2022-makes} to optimize the generation process.

\paragraph{CFs generation methods.} There exists several methods for generating CFs, but most of them are desigend for a specific LLM. The CFs generated by MICE~\cite{ross-etal-2021-explaining} are intended for debugging models, and not for data augmentation. Polyjuice~\cite{wu-etal-2021-polyjuice} requires specifying the type of edits that should be conducted, and the resulting CFs should be manually labeled. \cite{robeer-etal-2021-generating-realistic} DISCO~\cite{chen-etal-2023-disco} uses GPT-3's fill-in-the-blanks mode, which is unavailable in most open source LLMs and would require adapting them. CREST~\cite{treviso-etal-2023-crest} depends on a rationalizer module and the editor module is a masked LM that needs to be further trained. Instead, we decided to prompt LLMs to generate CFs by providing instructions and an example. We provide more details in Section~\ref{subsec:cfs_generation}.

\paragraph{LLMs for CFs generation}
\cite{li-etal-2024-prompting} investigated the strengths and weaknesses of LLMs as CFs generators. Additionally, they disclosed the factors that impact LLMs during CFs generation, including both intrinsic properties of LLMs and prompt design considerations. However, this study lacks intrinsic evaluation of CFs and omits  comparison with human-generated CFs. \citet{sachdeva-etal-2024-catfood} leverage LLMs to generate CFs for extractive question answering, showing that data augmentation with CFs improve OOD performance, and that this improvement correlates with the diversity of the generated CFs. 
Prior work by \citet{bhattacharjee2024llmguided} investigated the capability of GPT models in generating CFs for explanations by optimizing their prompts. However, their analysis was limited to the GPT family and did not consider downstream tasks or comparison with human-generated CFs.
In this work, we conduct a more comprehensive evaluation of LLMs on multiple aspects, including intrinsic metrics of CFs explanation quality and performance on downstream tasks. Furthermore, we compare the LLM-generated CFs against those produced by humans, and propose a novel approach to evaluate CFs using LLMs.

% both human evaluation and
%update work for counterfactual
% 2. Prompting methods 
% CoT
% https://aclanthology.org/2022.acl-long.556/

% capabilities of LLMs
%

\section{Conclusion}
\label{sec:conclusion}
%LLMs demonstrate value in generating CFs for both explanation and downstream tasks like data augmentation. In explanations, LLMs can produce CFs satisfying intrinsic metrics in sentiment analysis (SA), though they struggle with label flipping in Natural Language Inference (NLI) compared to humans. For data augmentation, some LLMs generate useful CFs for SA but fail to improve NLI tasks. Notably, the correlation between intrinsic metrics and accuracy in augmentation tasks suggests that generating fluent CFs might enhance performance of the classification. Finally, for counterfactual evaluation, GPT-4 appears as a strong candidate, while GPT-3.5 seems less suitable.

% Alternative
In this work, we investigated the use of various LLMs for CFs generation. Our results show that LLMs generate fluent CFs, but struggle to keep the induced changes minimal. Generating CFs for SA is less challenging than NLI and HS, where LLMs show weaknesses in generating CFs that change the original label. CFs from LLMs can replace human CFs for the purpose of data augmentation on SA and achieve similar performance, while on NLI and HS further improvements are needed. Further, our results suggest that CFs with minimal changes are essential for data augmentation.  % Qualitative analysis ?
We also showed that when asked to asses CFs, LLMs exhibit a strong bias towards agreeing with the provided label even if this label is incorrect. GPT4 appears to be more robust than GPT3.5 against this bias. 
%However, some failures still exist as shown in our qualitative analysis, indicating limitations of LLMs that warrant further investigation. 
Furthermore, we showed that GPT4 scores its own generations higher and that safety training might be one reason for this preference, i.e., GPT4 prefers its own generations, because they do not contain any harmful content. Future work should focus on (i) leveraging LLMs for higher quality NLI and HS CFs, which correctly change the label and keep changes minimal, (ii) assessing the evaluation abilities of LLMs in mislabeled data settings, and (iii) investigating the effects of safety training on LLMs as evaluators. %\TODO{update and mention HS}

%Furthermore, we showed that GPT4 prefers its own generations, giving them higher scores. Our analysis showed that one reason for this preference might be safety training, i.e., GPT4 prefers its own generations, because these generations do not contain any harmful content. 

% \clearpage
\section{Limitations}
%default hyperparameters for LLMs
%data leakage gpt3.4 and gpt4 ?

We used the default parameters for generating counterfactuals. Experimenting with different parameters might have a non-negligble effect on the results.
We included various LLMs in our experiments to be inclusive and be able to compare open-source and closed LLMs. However, these LLMs might have been exposed, during their training, to the data we use from~\cite{Kaushik-etal-2020-learning}. In this regard, the training data of most open-source and all closed-source LLMs remains unknown. In our qualitative analysis (see Section~\ref{subsec:qualitative}), we noticed that GPT4 generated a CF that is identical to a human CF from~\cite{Kaushik-etal-2020-learning}.


\bibliography{anthology,custom}
\bibliographystyle{acl_natbib}

\appendix

\section{Successful Generations}
\label{app:robustness}
%\subsection{Robustness Evaluation} 
Table~\ref{tab:robustness} shows how often LLMs successfully generated CFs, i.e., how often they adhered to the pre-defined template in the prompt.

\begin{table}[h]
    \centering
        \begin{tabular}{p{4.5cm}r}
        \toprule
         \textbf{Test split} &  \textbf{Success Rate}  \\
            \midrule

        \textbf{SA} & \\

        \midrule
    GPT3.5 &        100.00 \\
       GPT4 &         99.59 \\
  LLAMA2 7B &         98.98 \\
 LLAMA2 70B &         81.76 \\
 MISTRAL 7B &         84.22 \\
MISTRAL 56B &         94.67 \\

    \midrule
        \textbf{NLI}  & \\
    \midrule
\textbf{changed premise}  & \\\midrule
     GPT3.5 &        100.00 \\
       GPT4 &        100.00 \\
  LLAMA2 7B &         96.00 \\
 LLAMA2 70B &         98.00 \\
 MISTRAL 7B &         96.12 \\
MISTRAL 56B &         99.25 \\
\midrule
    \textbf{changed hypothesis} &  \\\midrule
    GPT3.5 &        100.00 \\
       GPT4 &        100.00 \\
  LLAMA2 7B &         99.62 \\
 LLAMA2 70B &         94.38 \\
 MISTRAL 7B &         94.38 \\
MISTRAL 56B &         98.25 \\
\midrule
    \textbf{HS} &  \\\midrule

     GPT3.5 &         63.21 \\
       GPT4 &         76.28 \\
 MISTRAL 7B &         80.44 \\
MISTRAL 56B &         81.44 \\
        \bottomrule
        \end{tabular}
    \caption{Success rate in generating CFs. We consider generations that do not adhere to the pre-defined template in the prompt as failed generations. }
    \label{tab:robustness}
\end{table}

\section{Further Analysis}
\label{app:further_analysis}
Table~\ref{tab:copypaste} shows the percentage of copy/paste examples in NLI CFs. 0

\begin{table}
    \begin{tabular}{llr}
\toprule
        \textbf{LLM} &      \textbf{changed part} &  \textbf{percentage}  \\
\midrule
    Crowd &    premise &        0.00 \\
      Crowd & hypothesis &        0.00 \\
     GPT3.5 &    premise &        1.63 \\
     GPT3.5 & hypothesis &        4.27 \\
       GPT4 &    premise &        4.14 \\
       GPT4 & hypothesis &        2.25 \\
  LLAMA2 7B &    premise &        3.00 \\
  LLAMA2 7B & hypothesis &        2.26 \\
 LLAMA2 70B &    premise &        2.04 \\
 LLAMA2 70B & hypothesis &        1.06 \\
 MISTRAL 7B &    premise &        0.91 \\
 MISTRAL 7B & hypothesis &        1.59 \\
MISTRAL 56B &    premise &        0.63 \\
MISTRAL 56B & hypothesis &        1.40 \\

\bottomrule
\end{tabular}
\caption{Percentage of CFs for each LLM where the CFs were a copy of the premise/hypothesis.}
\label{tab:copypaste}
\end{table}

\section{Hyperparameter Tuning}
\label{app:hptuning}
In order to evaluate how beneficial are the generated counterfactuals when used for data augmentation, we train several models with and without the generated counterfactuals. When training these, we conduct hyperparameter optimization, and choose the best model based on its performance on a validation set. We considers the following hyperparameters: batch size = [16, 32, 64], training epochs = [2,3,4,5], learning rate$~U(1e-5, 5e-5)$, weight decay $U(0, 0.3)$. In our experiments, we use HuggingFace~\cite{wolf-etal-2020-transformers} for training and evaluation of perplexity. For distance, we use spaCy~\footnote{\url{https://github.com/explosion/spaCy}}.

\section{Examples}
\label{app:examples}

We show categorize examples in Table~\ref{tab:categorized_examples}. Some correct and incorrect evaluation examples are depicted in Figure~\ref{tab:eval_examples}. An example of LLMs evaluation on an SA example is shown in Figure~\ref{fig:llm_eval:example}.% An example for NLI is shown in 

\begin{table*}

    \centering
    \tiny
\begin{tabular}{p{2cm}p{2cm}p{2cm}llllllll}
\toprule
Premise &    Hypothesis &  Counterfactual & Original Label & Target Label & FL & UA & RS &         LLM &     Changed Part \\

\midrule
Correct &                                                                  &                                                                                                                         &                &                &            &                  &              &            &                    \\ \midrule
                                        An older man sifts water through his fingers as a young toddler watches from inside the water. &                                          A man is near a toddler &                                An older man sifts water through his fingers as a young toddler is far from any toddler. &     entailment &  contradiction &        1.0 &              1.0 &          1.0 &       GPT4 &    revised\_premise \\
%A photographer and assistant take a photo of a man in a suit and a woman in a dress in front of large white columns and a yellow wall. &                     The photographer is taking a photo of a dog. &                                      The photographer takes a photo of a dog and a man in a suit and a woman in a dress &  contradiction &     entailment &        4.0 &              4.0 &          4.0 & llama2\_70b & revised\_hypothesis \\ 
\midrule
Incorrect  &                                                                   &                                                                                                                         &                &                &            &                  &              &            &                    \\ \midrule
                                                   At least eight people are sitting at a table in a building that overlooks the city. &                                      People sitting on the floor & At least eight people are sitting at a table in a building that overlooks the city. People can be sitting on the floor. &  contradiction &     entailment &        4.0 &              4.0 &          4.0 &    Mistral 7B &    revised\_premise \\
%                               A boy dressed for summer in a green shirt and kahki shorts extends food to a reindeer in a petting zoo. & A boy alien dressed for summer in a green shirt and kahki shorts &                                                       A girl alien dressed for summer in a green shirt and kahki shorts &     entailment &  contradiction &        1.0 &              1.0 &          1.0 & llama2\_70b & revised\_hypothesis \\
\bottomrule
\end{tabular}
\caption{Correct and incorrect evaluation examples from GPT4. }
\label{tab:eval_examples}
\end{table*}

%\input{src/examples/sent}
%\input{src/examples/nli_premise}
%\input{src/examples/nli_hypo}

%todo: need to decrease the font size
\begin{figure*}

\setlength\fboxsep{7pt}
\fbox{%
\parbox{0.95\linewidth}{%
\footnotesize
\begin{courier}
Text 1, Factual Label = Negative: Waiting to go inside the theathre with tickets in my hand, I expected an interesting sci-fi fantasy movie which could finally feed my appetite of movies regarding robot-technology, instead I went disappointed by each aspect of it, once more proving that stunning special effects can't help a boring plot, which by my opinion was the worse in this year. Acting in this movie also dissatisfied me, Will Smith didn't show anything new in this movie, yet I never saw his acting to change since "Men In Black" which was his only success by my opinion. He had to retire since than, not spoiling his name with titles like "I,Robot" and "Men In Black 2". 4/10\\

Text 2, Target Label = Positive: Waiting to go inside the theatre with tickets in my hand, I saw an interesting sci-fi fantasy movie which could finally feed my appetite for movies regarding robot-technology, I wasn't  disappointed by any aspect of it, once more proving that stunning special effects can help the plot, which by my opinion was the best this year. Acting in this movie satisfied me, Will Smith didn't show anything new in this movie, yet I never saw his acting to change since "Men In Black" which was one of his biggest success in my opinion. ". 8/10

Answer:\\
1. 4\\
2. 2\\
3. 2\\

Explanation for 3: The overall sentiment toward the movie has been successfully inverted from negative to positive. However, the remark about Will Smith's acting not changing since "Men In Black" remained unchanged, which conflicts with the positive sentiment of the rest of the review. This inconsistency may cause readers to perceive the review as somewhat unrealistic because the critique of the actor’s performance persisting in an otherwise positive review is unusual.

4. 3 \\
5. 2 \\
6. 2 \\

\end{courier}
}}
\caption{GPT-4 evaluation for a crowd-generated text in a sentiment analysis task, with a good explanation, as it can identify the negative aspects in the revised review.}
\label{fig:llm_eval:example}
\end{figure*}

%\section{Data}
%\label{app:data}
%\input{src/app_data}

\section{Prompt}
\label{app:prompt}
Prompts for generating SA CFs and NLI CFs are shown in Figure~\ref{fig:prompt:gen:sent} and~\ref{fig:prompt:gen:nli} respectively. Prompts for evaluating SA CFs and NLI CFs are shown in Figure~\ref{fig:llm_eval:sa} and ~\ref{fig:llm_eval:nli} respectively. 

\begin{figure*}
\setlength\fboxsep{7pt}
\fbox{%
\parbox{0.95\linewidth}{%
\begin{courier}
Request: Given a piece of text with the original sentiment in the form of "Sentiment: Text". Change the text with minimal edits to get the target sentiment from the original sentiment. Do not make any unneccesary changes.\\
(Original text)\\
Positive: This is an excellent film!Tom Hanks and Paul Newman performed great!I was really surprised when Newman was beating on his son!That was a great scene and the shooting scenes were staged good.I was very surprised about the end.Rent this film today as it is one of Tom Hanks' best!\\
Target: Negative\\
\\
Step 1: Identify phrases, words leading to the Positive sentiment: \\
'an excellent',\\
'great!I',\\
'a great',\\
'good.I',\\
'end.Rent ',\\
'today',\\
'best!',\\
\\
Step 2: Change these phrases, words, sentences to Negative sentiment with minimal changes: \\
'an excellent' to 'a terrible'\\
'great!I' to 'badly!I'\\
'a great' to 'an awful'\\
'good.I' to 'badly.I'\\
'end.Rent ' to 'end.Don't rent'\\
'today' to ''\\
'best!' to 'worst!'\\
\\
Step 3: replace the phrases, words, sentences from step 1 in the original text by the phrases, words, sentences in step 2:\\
Negative: This is a terrible film!Tom Hanks and Paul Newman performed badly!I was really surprised when Newman was beating on his son!That was an awful scene and the shooting scenes were staged badly.I was very surprised about the end.Don't rent this film as it is one of Tom Hanks' worst!\\
\\
Request: Given a piece of text with the original sentiment in the form of "Sentiment: Text". Change the text with minimal edits to get the target sentiment from the original sentiment. Do not make any unnecessary changes.\\
(Original text)\\
Positive: Well, sorry for the mistake on the one line summary.......Run people, run..to your nearest movie store, that is! This movie is an fabulous!! Imagine! Gary Busey in another low budget movie, with an incredibly funny scenario...isn't that a dream? No (well yes), it is Plato's run...........I give it ****  out of *****. \\
Target: Negative  \\

\end{courier}
}}
\caption{Prompt for CFs generation - SA}
\label{fig:prompt:gen:sent}
\end{figure*}
\begin{figure*}
\setlength\fboxsep{7pt}
\fbox{%
\parbox{0.95\linewidth}{%
\begin{courier}

Given two sentences (premise and hypothesis) and their original relationship, determine whether they entail, contradict, or are neutral to each other. Change the sentence2 with minimal edits to achieve the target relation from the original one. Do not make any unnecessary changes. For example:\\
Original relation: entailment\\
Two original sentences: Brown-haired woman talking to man with backpack. A woman is talking to a man\\
Target relation: neutral\\
Target sentence: sentence2\\
Step 1: Identify phrases, words in the sentence2 leading to the entailment relation: \\
'man',\\
\\
Step 2: Change these phrases, words to get neutral relation with minimal changes: \\
'man' to 'student.'\\
\\
Step 3: replace the phrases, words from step 1 in the original text by the phrases, words, sentences in step 2:\\
(Edited sentence2): A woman is talking to a student.\\
\#\#\#\#\#End Example\#\#\#\#\\
Request: Given two sentences (premise and hypothesis) and their original relationship, determine whether they entail, contradict, or are neutral to each other. Change the sentence2 with minimal edits to achieve the neutral relation from the original one. Do not make any unnecessary changes. Do not add anything else.\\
Original relation: entailment\\
Two original sentences: A blond woman speaking to a brunette woman with her arms crossed. A woman is talking to another woman.\\
Target relation: neutral\\
Target sentence: sentence2

\end{courier}
}}
\caption{Prompt for CFs generation - NLI}
\label{fig:prompt:gen:nli}
\end{figure*}

% \input{src/prompts/llm_eval_sent}
%todo: need to decrease the font size
\begin{figure*}

\setlength\fboxsep{7pt}
\fbox{%
\parbox{0.95\linewidth}{%
\footnotesize
\begin{courier}
Evaluating Counterfactuals

Natural Language Inference (NLI) is a fundamental task in natural language processing (NLP) that involves determining the relationship between two pieces of text: a premise and a hypothesis. The relation between the premise and the hypothesis is described using three different labels:

Entailment: if the hypothesis is definitely true given the premise. \\
Example for Entailment: \\
Premise: A soccer game with multiple males playing. \\
Hypothesis: Some men are playing a sport. \\
Label: Entailment

Neutral: if the hypothesis might be true given the premise. \\
Example for Neutral: \\
Premise: An older and younger man smiling. \\
Hypothesis: Two men are smiling and laughing at the cats playing on the floor. \\
Label: Neutral

Contradiction: if the hypothesis is definitely false given the premise. \\
Example for Contradiction: \\
Premise: A man inspects the uniform of a figure in some East Asian country. \\
Hypothesis: The man is sleeping \\
Label: Contradiction

Purpose of the Evaluation: \\
This evaluation aims to assess the quality of counterfactual texts that were generated by different methods. A counterfactual text is an alternative version of a text designed to change the label of the original (factual) instance while maintaining high text quality.

Task Description: \\
You will receive two instances. Each instance consists of two sentences: a premise and a hypothesis. Each instance can be classified with one of the three aforementioned labels (Entailment, Neutral, Contradiction).

Factual instance (Instance 1): An instance and its factual label. \\
Counterfactual instance (Instance 2): A modified version of the factual instance designed to express a different label, i.e., match the target label.

Read the two texts and answer the questions below:

Instance 1: \\
Premise: {} \\
Hypothesis: {} \\
Factual Label: {}

Instance 2: \\
Premise: {} \\
Hypothesis: {} \\
Target Label: {}

1. To which extent do you agree that Instance 2 has the label {}? \\
   4-totally agree, 3-partially agree, 2-partially disagree, 1-totally disagree

2. Are there any unnecessary changes (removals, additions, replacements of words) in the counterfactual text (Instance 2) that do not contribute to changing the original factual label to the target label? \\
   4-no unnecessary changes, 3-few unnecessary changes, 2-many unnecessary changes, 1-significant number of unnecessary changes

3. How realistic is Instance 2? A realistic instance would not include any imaginary actions/items. \\
   4-very realistic, 3-partially realistic, 2-partially unrealistic, 1-very unrealistic

If you think it is (highly/partially) unrealistic, please provide a brief explanation.

Your evaluation for the provided counterfactual text: \\
1. Please provide a number only \\
2. Please provide a number only \\
3. Please provide a number only

\end{courier}
}}
\caption{Prompt for CFs Evaluation - NLI}
\label{fig:llm_eval:nli}
\end{figure*}
%todo: need to decrease the font size
\begin{figure*}

\setlength\fboxsep{7pt}
\fbox{%
\parbox{0.95\linewidth}{%
\footnotesize
\begin{courier}
Evaluating Counterfactuals\\

Texts can be classified into different categories, e.g., positive vs. negative sentiment. In this case the sentiment (positive/negative) is called the `label`. A counterfactual text is an alternative version of a text designed to change the label of the original (factual) text while maintaining high text quality.\\

Purpose of the Evaluation: \\
This evaluation aims to assess the quality of counterfactual texts that were generated by different methods. \\

Task Description: \\
You will receive two texts. Each text can either express a positive or a negative sentiment. \\
Factual Text (Text 1): A movie review with its (ground truth) factual label.\\
Counterfactual Text (Text 2): A modified version of the movie review designed to express the opposite sentiment, i.e., match the target label.\\

A simple example:
Text 1, Factual Label = Negative: This movie is very bad.\\
Text 2, Target Label = Positive:  This movie is great. \\

Read the two texts and answer the questions below:\\

Text 1, Factual Label = {}: {}\\

Text 2, Target Label = {}: {}\\

1. To which extent do you agree that Text 2 has the label {}? \\
4-totally agree, 3-partially agree, 2-partially disagree, 1-totally disagree\\

2. Are there any unnecessary changes (removals, additions, replacements of words) in the counterfactual text (Text 2) that do not contribute to changing the original factual label to the target label?\\
4-no unnecessary changes, 3-few unnecessary changes, 2-many unnecessary changes, 1-significant number of unnecessary changes\\

3. How realistic is Text 2? A realistic movie review would not read strange in any way on a movie review website. \\
4- very realistic, 3-partially realistic, 2-partially unrealistic, 1-very unrealistic\\

If you think it is (highly/partially) unrealistic, please provide a brief explanation. \\

Additionally, assess the following aspects of the counterfactual text:\\
4. Grammaticality:  how would you rate the grammatical accuracy of text 2?\\
4-Definitely correct, 3-Somewhat correct, 2-Somewhat incorrect, 1-Definitely incorrect\\
5. Cohesiveness: How well do the sentences in the text 2 fit together? \\
4-Highly cohesive, 3-Reasonably cohesive, 2-Somewhat disjointed, 1-Very poorly fit together\\
6. Likability: how likely are you to vote for text 2 on the movie review site?\\
4-Definitely would vote, 3-Likely to vote, 2-Unlikely to vote, 1-Definitely would not vote\\

Your evaluation for the provided counterfactual text:\\
1. (Please provide a number only)\\
2. (Please provide a number only)\\
3. (Please provide a number only)\\
4. (Please provide a number only)\\
5. (Please provide a number only)\\
6. (Please provide a number only)

\end{courier}
}}
\caption{Prompt for CFs Evaluation - SA}
\label{fig:llm_eval:sa}
\end{figure*}

\end{document}